\documentclass{article}

\usepackage[preprint,nonatbib]{neurips_2022}

\usepackage[utf8]{inputenc} 
\usepackage[T1]{fontenc}    
\usepackage{hyperref}       
\usepackage{url}            
\usepackage{booktabs}       
\usepackage{amsfonts}       
\usepackage{nicefrac}       
\usepackage{microtype}      
\usepackage{xcolor}         
\usepackage{pgfplots}
\usepackage{physics}        
\pgfplotsset{width=7cm,compat=1.8}

\usepackage{graphicx}
\usepackage{amsmath}
\usepackage{amssymb}
\usepackage{booktabs}
\usepackage{algorithm2e}

\usepackage{pgfplots}
\usepackage{pgfplotstable}

\usepackage{tikz}
\usetikzlibrary{arrows.meta,
                matrix,
                calc,
                positioning}

\usepackage{subcaption}
\pgfplotstableread{
x         y    y-max  y-min
SGD+bs=25 26.536 2.986000000000004 2.6140000000000043
SGD+bs=50 26.614000000000004 3.1140000000000043 2.7759999999999962
SGD+bs=100 26.638000000000005 3.1780000000000115 2.721999999999994
}{\sgddata}

\pgfplotstableread{
x         y    y-max  y-min
Adam+bs=25 25.92 4.489999999999995 3.450000000000003
Adam+bs=50 26.510000000000005 3.760000000000005 4.719999999999999
Adam+bs=100 27.010000000000005 4.890000000000001 1.9499999999999886
}{\adamdata}

\pgfplotstableread{
x         y    y-max  y-min
DBA+SGD+VarianceNorm 3.019999999999996 0.11999999999999034 0.13000000000000966
DBA+SGD+GradientNorm 2.662000000000006 0.2220000000000084 0.367999999999995
}{\dbasgddata}

\pgfplotstableread{
x         y    y-max  y-min
DBA+Adam+VarianceNorm 9.069999999999993 0.35999999999999943 0.480000000000004
DBA+Adam+GradientNorm 8.337999999999994 0.14799999999999613 0.24200000000000443
}{\dbaadamdata}

\pgfplotstableread{
x         y    y-max  y-min
SGD 26.536 2.986000000000004 2.6140000000000043
Adam 25.92 4.489999999999995 3.450000000000003
DBA+SGD 2.662000000000006 0.2220000000000084 0.367999999999995
DBA+Adam 8.337999999999994 0.14799999999999613 0.24200000000000443
}{\totaldba}

\title{Dynamic Batch Adaptation}

\author{%
    Cristian Simionescu \\
    \footnotesize{Faculty of Computer Science}\\
    \footnotesize{"Alexandru Ioan Cuza" University}\\
    \scriptsize{\texttt{cristian@nexusmedia.ro}} \\
    \And
    George Stoica \\
    \footnotesize{Faculty of Computer Science}\\
    \footnotesize{"Alexandru Ioan Cuza" University}\\
    \scriptsize{\texttt{sgeorge.sstoica99@gmail.com}}\\
    \And
    Robert Herscovici\\
    \footnotesize{Independent Researcher}\\
    \scriptsize{\texttt{hhroberthdaniel@gmail.com}} \\
}

\begin{document}

\maketitle


\begin{abstract}
    Current deep learning adaptive optimizer methods adjust the step magnitude of parameter updates by altering the effective learning rate used by each parameter.
    Motivated by the known inverse relation between batch size and learning rate on update step magnitudes, we introduce a novel training procedure that dynamically decides the dimension and the composition of the current update step.
    Our procedure, Dynamic Batch Adaptation (DBA) analyzes the gradients of every sample and selects the subset that best improves certain metrics such as gradient variance for each layer of the network. 
    We present results showing DBA significantly improves the speed of model convergence.
    Additionally, we find that DBA produces an increased improvement over standard optimizers when used in data scarce conditions where, in addition to convergence speed, it also significantly improves model generalization, managing to train a network with a single fully connected hidden layer using only 1\% of the MNIST dataset to reach 97.79\% test accuracy.
    In an even more extreme scenario, it manages to reach 97.44\% test accuracy using only 10 samples per class. 
    These results represent a relative error rate reduction of 81.78\% and 88.07\% respectively, compared to the standard optimizers, Stochastic Gradient Descent (SGD) and Adam.
\end{abstract}

\section{Introduction}


The convergence process of a neural network is heavily influenced by which optimizer is used during training. 
An optimizer that can reach convergence quickly enables the model to achieve better generalization in less parameter updates, therefore reducing computational costs and allowing for longer and more complex experiments. 
Modern optimizers, such as Adam \cite{kingma2014adam}, adjust the learning rate of parameters in order to increase the speed of convergence. 
Adam does this by taking into consideration the exponential running mean of the previous gradients, maintaining a separate learning rate for each weight.  

The inverse relation between batch size and learning rate has been shown before by ~\cite{smith2017don}. 
They demonstrate that increasing the batch size leads to a similar result as decreasing the learning rate. 
Using a higher batch size comes with the benefit of better hardware utilization and enhanced parallelization. 
Moreover, fewer update steps are required in order to achieve convergence and produces similar results as a method which adjusts the learning rate.
As a consequence, since it has been shown that we can use a constant learning rate and modify the batch size, several studies have attempted to use this characteristic to develop new training techniques.
These previous works have mainly focused on changing the batch size using schedulers \cite{devarakonda2017adabatch}, \cite{khan2020adadiffgrad} or by analyzing data from the training history \cite{alfarraadaptive}.

Our proposed solution is to dynamically select which gradients are used to calculate the update for each layer, while also providing a method of adjusting the batch size at the end of each epoch. These gradients are chosen from the current batch which was forward passed through the model.
Weight updates are characterized by having two important components, their direction and magnitude. 
For a fixed batch, changing the learning rate results in the magnitude of the update being adjusted, however, for a fixed learning rate, modifying the number of samples included in the batch allows us to impact both the magnitude and the direction.
Previous works which used the change of batch size as a method of adjusting the magnitude of the update gradient didn't focus their research on the possibility of adjusting the direction, which we consider to bring more benefits to the neural network's training process.
Therefore, our approach consists of selecting sample gradients that are to be included in the update gradient in order to improve both it's direction and magnitude.
The adjustment to the batch size is done in order to allow a qualitative selection process while also not wasting computational resources if only a small fraction of the sample gradients from each batch are included in the parameter update.

In Section \ref{other-works} we present the research done in other works which study the effect of batch size upon the convergence of the network.
In Section \ref{our-work} we introduce our optimization algorithm which selects the samples that are to be included in an update and modifies the batch size.
Next in Section \ref{experiments} we discuss the experiments and the results obtained using our optimization algorithm, in Section \ref{limitations} we talk about limitations and finally, the conclusion in Section \ref{conclusion}.

\section{Related work}
\label{other-works}

Improving model convergence by adjusting learning rates or adapting the batch size has been attempted before. 
Previous works in literature studied the effect of the batch size on the convergence process of neural networks and possible ways in which the training speed can be improved with regards to batch size.
In \cite{li2014efficient} the authors analyze whether the convergence rate decreases with the increase in batch size and show that bigger batch sizes do not hinder convergence.
Since \cite{smith2017don} showed that similar results can be obtained by both approaches, dynamic adaptation of batches has garnered attention since in addition to model convergence, it also has the potential to address issues related to hardware utilization and computation parallelism. 

Recent works by \cite{devarakonda2017adabatch} and \cite{khan2020adadiffgrad}, propose the use of schedulers which increase the batch size at set points during training similarly to the popularly equivalent technique applied to learning rate schedulers.
These methods bring the benefit of better convergence and hardware utilization but require an initial exploration to identify suitable milestones. 
Alternatively, \cite{lederrey2021estimation} suggests increasing the batch size when the model hits a plateau which is less sensitive to hyperparameter choices but has the potential of waiting too long before making a change.

Other authors proposed that batch size should be increased when fulfilling a certain criteria, namely a reduction in the loss of the model \cite{liu2019accelerate} or based on the variance of the gradients for the current epoch \cite{balles2016coupling}, while \cite{gao2020balancing} suggests updating the batch size by a static amount when the convergence rate decreases under a certain threshold. 
In \cite{alfarraadaptive}, authors propose a practical algorithm to approximate an optimal batch size by storing information regarding previously used gradients in order to decide a batch size for the current iteration. 

Compared with previous research, our proposed method changes not only the batch size but also its composition, by using only a subset of the samples from current batch in order to perform the update of the weights. 
Another important distinction of our works is that we do the selection of the samples for each layer of the network separately.

\section{Dynamic Batch Adaptation}
\label{our-work}
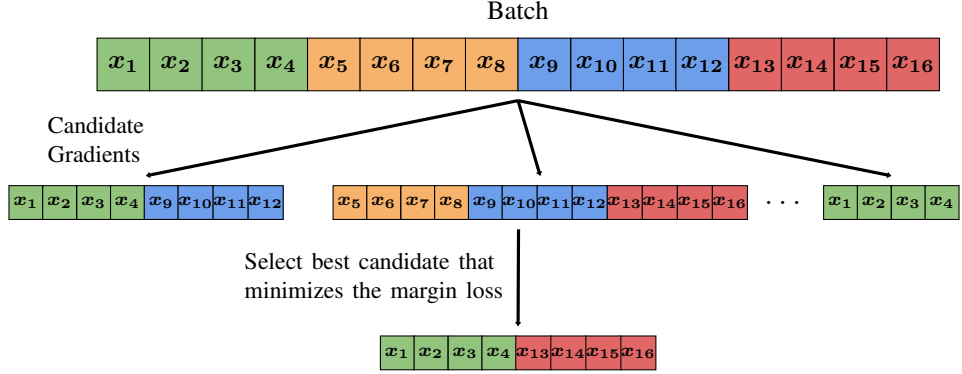
\begin{figure}
    \centering
    
    \begin{tikzpicture}[
        node distance = 4mm,
          MTRX/.style = {matrix of nodes,
                          nodes in empty cells,
                         nodes={draw, minimum size=7.0mm, anchor=center,
                                inner sep=0pt, outer sep=0pt},
                         column sep=-\pgflinewidth,
                         row sep=2mm},
          MTRXSmall/.style = {matrix of nodes,font=\scriptsize,
                 nodes={draw, minimum size=4.5mm, anchor=center,
                        inner sep=0pt, outer sep=0pt},
                 column sep=-\pgflinewidth,
                 row sep=2mm},
                                ]
                                
        \node[text width=1.5cm] at (-5.5,-1) {\small Candidate Gradients};  
                                  
        \matrix  (m1) [MTRX,
                        row 1/.append,
                       label=above: Batch,
                       ] 
        {
        |[draw,fill={rgb,255:red,147; green,196; blue,125}]|$\boldsymbol{x_1}$ & |[draw,fill={rgb,255:red,147; green,196; blue,125}]|$\boldsymbol{x_2}$ & |[draw,fill={rgb,255:red,147; green,196; blue,125}]|$\boldsymbol{x_3}$ & |[draw,fill={rgb,255:red,147; green,196; blue,125}]|$\boldsymbol{x_4}$ & |[draw,fill={rgb,255:red,247; green,179; blue,107}]|$\boldsymbol{x_5}$ & |[draw,fill={rgb,255:red,247; green,179; blue,107}]|$\boldsymbol{x_6}$ & |[draw,fill={rgb,255:red,247; green,179; blue,107}]|$\boldsymbol{x_7}$ & |[draw,fill={rgb,255:red,247; green,179; blue,107}]|$\boldsymbol{x_8}$ & |[draw,fill={rgb,255:red,109; green,158; blue,235}]|$\boldsymbol{x_9}$ & |[draw,fill={rgb,255:red,109; green,158; blue,235}]|$\boldsymbol{x_{10}}$ & |[draw,fill={rgb,255:red,109; green,158; blue,235}]|$\boldsymbol{x_{11}}$ & |[draw,fill={rgb,255:red,109; green,158; blue,235}]|$\boldsymbol{x_{12}}$ & |[draw,fill={rgb,255:red,225; green,103; blue,102}]|$\boldsymbol{x_{13}}$ & |[draw,fill={rgb,255:red,225; green,103; blue,102}]|$\boldsymbol{x_{14}}$ & |[draw,fill={rgb,255:red,225; green,103; blue,102}]|$\boldsymbol{x_{15}}$ & |[draw,fill={rgb,255:red,225; green,103; blue,102}]|$\boldsymbol{x_{16}}$\\
        };

        \matrix  (m2) [MTRXSmall, below left=1.0cm and -2.74cm of m1
                      ]
        {
          |[draw,fill={rgb,255:red,147; green,196; blue,125}]|$\boldsymbol{x_1}$ & |[draw,fill={rgb,255:red,147; green,196; blue,125}]|$\boldsymbol{x_2}$ & |[draw,fill={rgb,255:red,147; green,196; blue,125}]|$\boldsymbol{x_3}$ & |[draw,fill={rgb,255:red,147; green,196; blue,125}]|$\boldsymbol{x_4}$ & 
          |[draw,fill={rgb,255:red,109; green,158; blue,235}]|$\boldsymbol{x_9}$ & |[draw,fill={rgb,255:red,109; green,158; blue,235}]|$\boldsymbol{x_{10}}$ & |[draw,fill={rgb,255:red,109; green,158; blue,235}]|$\boldsymbol{x_{11}}$ & |[draw,fill={rgb,255:red,109; green,158; blue,235}]|$\boldsymbol{x_{12}}$ \\
        };
        \matrix  (m3) [MTRXSmall, right=of m2
                      ]
        {
          |[draw,fill={rgb,255:red,247; green,179; blue,107}]|$\boldsymbol{x_5}$ & |[draw,fill={rgb,255:red,247; green,179; blue,107}]|$\boldsymbol{x_6}$ & |[draw,fill={rgb,255:red,247; green,179; blue,107}]|$\boldsymbol{x_7}$ & |[draw,fill={rgb,255:red,247; green,179; blue,107}]|$\boldsymbol{x_8}$ & |[draw,fill={rgb,255:red,109; green,158; blue,235}]|$\boldsymbol{x_9}$ & |[draw,fill={rgb,255:red,109; green,158; blue,235}]|$\boldsymbol{x_{10}}$ & |[draw,fill={rgb,255:red,109; green,158; blue,235}]|$\boldsymbol{x_{11}}$ & |[draw,fill={rgb,255:red,109; green,158; blue,235}]|$\boldsymbol{x_{12}}$ & |[draw,fill={rgb,255:red,225; green,103; blue,102}]|$\boldsymbol{x_{13}}$ & |[draw,fill={rgb,255:red,225; green,103; blue,102}]|$\boldsymbol{x_{14}}$ & |[draw,fill={rgb,255:red,225; green,103; blue,102}]|$\boldsymbol{x_{15}}$ & |[draw,fill={rgb,255:red,225; green,103; blue,102}]|$\boldsymbol{x_{16}}$\\
        };
        
        \matrix  (m4) [MTRXSmall, right=0.75cm of m3
                      ]
        {
            |[draw,fill={rgb,255:red,147; green,196; blue,125}]|$\boldsymbol{x_1}$ & |[draw,fill={rgb,255:red,147; green,196; blue,125}]|$\boldsymbol{x_2}$ & |[draw,fill={rgb,255:red,147; green,196; blue,125}]|$\boldsymbol{x_3}$ & |[draw,fill={rgb,255:red,147; green,196; blue,125}]|$\boldsymbol{x_4}$ \\
        };

        \matrix  (m5) [MTRXSmall, below=3.0cm of m1
                      ]
        {
            |[draw,fill={rgb,255:red,147; green,196; blue,125}]|$\boldsymbol{x_1}$ & |[draw,fill={rgb,255:red,147; green,196; blue,125}]|$\boldsymbol{x_2}$ & |[draw,fill={rgb,255:red,147; green,196; blue,125}]|$\boldsymbol{x_3}$ & |[draw,fill={rgb,255:red,147; green,196; blue,125}]|$\boldsymbol{x_4}$ & |[draw,fill={rgb,255:red,225; green,103; blue,102}]|$\boldsymbol{x_{13}}$ & |[draw,fill={rgb,255:red,225; green,103; blue,102}]|$\boldsymbol{x_{14}}$ & |[draw,fill={rgb,255:red,225; green,103; blue,102}]|$\boldsymbol{x_{15}}$ & |[draw,fill={rgb,255:red,225; green,103; blue,102}]|$\boldsymbol{x_{16}}$\\
        };                        
        
        \draw[->, -{Implies[]}, very thick] ($(m3.south) - (0.82em, 0)$) -- node [text width=3.5cm,midway,left ] {\small Select best candidate that minimizes the margin loss}  ($(m5.north) - (0, 0)$) ;

        \draw[->, -{Implies[]}, very thick] (m1.south) -- (m2.north);
        \draw[->, -{Implies[]}, very thick] (m1.south) -- (m3.north);
        \draw[->, -{Implies[]}, very thick] (m1.south) -- (m4.north);
        \path (m3) -- (m4) node [font=\large, midway, sloped] {$\dots$};
    \end{tikzpicture}

    \caption{Gradient Subset Selection}
    \label{fig:candidate_gradient}
\end{figure}

Our algorithm, in its general form, is a wrapper on top of any other optimizer and it works by processing the gradients for each layer before they are applied to the weights.
More exactly, for each layer we compute the gradient of every sample individually, and then select only a subset of them which is chosen by minimizing a margin-based loss between a \textit{gradient metric} and a \textit{model metric}.

We refer to \textit{model metric} as being an indicator for the model's current performance which we measure by the model loss.
The \textit{gradient metric} is a function, such as a norm, calculated from a set of per-sample gradients for a given network layer. 
However, instead of composing this set from individual samples, we group multiple of them in \textit{selection strides}, allowing for better performance and stability. 
We define \textit{selection stride} as being a subset of $s$ consecutive samples within a batch. 
In a batch of size $b$, samples $x_{0}$, $x_{1}$ \dots $x_{s-1}$ belong to the first \textit{selection stride}, samples $x_{s}$, $x_{s+1}$ \dots $x_{2s -1}$ belong to the second \textit{selection stride}, and so on.
The general formula for a selection stride is:
\begin{equation}
    \{x_{i \cdot s}, x_{i \cdot s + 1} \dots x_{i \cdot (s + 1) - 1}\} \in SelectionStride_{i}, \forall i \in \left[0, \left\lceil \frac{b}{s} \right\rceil  - 1 \right]
\end{equation}

These \textit{selection strides} are used to compose candidate gradients and select the one that best minimizes the margin-based loss function. 
This is exemplified in Fig. ~\ref{fig:candidate_gradient} where a batch of 16 samples is divided into 4 selection strides from which multiple candidate gradients are constructed and evaluated. 
The candidate which is the best fit for our criterion is used in the weight update.

Dynamic Batch Adaptation (DBA) is applied for every batch in a training epoch, for any layer that has update gradients.
Notice that since every layer has distinct gradients, different gradient candidates can be selected for each of them.
The chosen \textit{selection strides} are averaged in order to compute the gradient on the current layer.
After the gradient has been replaced on each layer of the neural network, the update of the weights is performed as usual. A pseudo-code description of how DBA processes the per-sample gradients after they have been back-propagated can be seen at Algorithm ~\ref{alg:step}.
At the end of each epoch, using information about how many samples were included into the actual update, the batch size for the next epoch is adjusted accordingly. 

\begin{algorithm}
\caption{Optimizer step}\label{alg:step}
\DontPrintSemicolon
  \SetKwFunction{FMain}{DBA::step}
  \SetKwProg{Fn}{Procedure}{:}{}
  \Fn{\FMain{$loss$}}{
        $modelMetric \gets ModelMetric(loss)$ \\
        \For{$layer \in model.layers$}{
            $selectionStrides \gets SplitStrides(layer.gradSamples, strideSize)$ \\
            $chosen \gets SelectionStrategy(selectionStrides)$ \\
            $layer.gradient \gets Mean(chosen)$
        }
        optimizer.step($loss$)
  }
\end{algorithm}

\subsection{Model metric}

Our intent with this metric is to introduce a mechanism that is tied to the model performance in order to adapt the gradient selection algorithm to the current state of the model. 
This was desired in order to handle multiple stages during training.
In the beginning, the model should make large updates, while later in the training process, when the network has already learned a good set of weights, further updates should not make large changes.
We have experimented with using constant functions, however they offer no information about the current state of the model and are very sensitive to the chosen constant value.
The model metric we have have used is the loss of the model. 

We consider that other metrics can be developed to fulfil this role, but for the rest of the paper we will use the model loss as the \textit{model metric} for our algorithm.
The loss is computed for each batch, therefore it offers us an approximation of classification proficiency of the model for the current samples.

\subsection{Gradient metric}

Using the \textit{model metric} calculated for a given batch loss, we are able to guide our selection by calculating the \textit{gradient metric} of candidate gradients.
Normally, the mean of the batch loss is back-propagated to every layer of the model.
However, for our method, we need to back-propagate gradients for each of the samples.
This technique has notably been extensively used in the field of Differential Privacy \cite{yousefpour2021opacus}.
After the initial gradient calculations, we iterate through each layer and select subsets of individual gradients that minimize our margin-based loss between \textit{model metric} and \textit{gradient metric}. 
A subset of samples, represented by one or more  \textit{selection strides} from the current batch, receives a score calculated using the \textit{gradient metric}. 
In our experiments, we used two different \textit{gradient metrics}, the norm of the mean of the gradients for the selected samples, and the norm of the variance of the gradients for the selected samples. The grad metric can use one of these two formulations:

\begin{equation}
\label{grad-metric}
    \text{Gradient Norm}=\norm{\frac{\sum_{j = 0}^{n - 1}\delta_{i_{j}} }{n}}_{2}, \text{ Variance Norm}=\norm{\frac{\sum_{j = 0}^{n - 1}(\delta_{i_{j}} - \overline{\delta_{i}})}{n - 1}}_{2}\\
\end{equation}
Where $\delta_{i_{j}}$ is the $j$th gradient sample from candidate gradient $i$ and $n$ being the size of the candidate gradient. 

We have observed that the norm of the gradient from one sample is usually bigger than the norm of the mean of the gradients from more samples. 
Generally, the more sample gradients we use, the norm of their mean is lower due to the fact that their directions cancel each other provided that there isn't a consensus between all samples. 
Initially, because the model has random weights, all samples would have gradients with big norms because they generate a large loss. 
As the model learns from the data, the norm of the gradients would decrease over time, hence samples which are classified correctly by the model would have a smaller corresponding norm than the ones which are classified incorrectly. 
Moreover, the norms are bigger for the final classification layer than for the intermediate ones, which is why we consider that it is important to compute a different update for each layer.
Using different samples for each layer can cause internal covariate shift \cite{ioffe2015batch} on models with more layers due to the fact that each layer is effectively trained on a slightly different distribution of samples, but we haven't observed this phenomena on the small model used for our experiments.

Noisy samples and outliers lead to gradients with bigger norms because the model is not able to classify them well.
Therefore, our selection strategy would avoid \textit{selection strides} with noisy samples, or would select them only if their direction is canceled by the other samples and the negative effects are ameliorated.
We consider this to potentially be the primary reason for our good results in data-scarce environments, since noisy data has a greater impact on the model.

The norm of the variance of the gradients for the selected sample is the second indicator we used in our experiments to compute the  \textit{gradient metric}.
This metric describes how close the direction of the gradients for each sample are from each other. 
A set of gradients with similar directions would have a small variance, and therefore a small norm, while gradients with opposing directions would have a big variance and a big resulting variance norm. 
Similarly, noisy samples increase the variance and the value of this metric, making it easier to identify them.
From our experiments we see that both the variance and the norm of the gradients produce similar results, but using the variance based metric tends to lead to slightly better ones.

\subsection{Margin-based loss}

Ideally our algorithm would cause the model to make large update steps at the beginning of training when the weights don't contain much information, and smaller steps towards the end of training when we only want to fine-tune the weights without destroying good feature representations it has already learned.
Because of this, we do not directly try to minimize the \textit{gradient metric} due to the fact that by doing so we would achieve a constant behaviour across the entire training phase. 
Hence, we minimize a function which takes as arguments the  \textit{model metric} and the \textit{gradient metric}. 

We experimented with multiple such functions, but we have decided to use a margin-based loss that minimizes the slope between the \textit{model metric} and the \textit{gradient metric}.
In order to do this we keep an exponential running mean for each of the two metrics, and we try to choose \textit{selection strides} such that the \textit{gradient metric} minimizes the slope as follows:
\begin{equation}
\label{argmin}
    arg \min_{X \subseteq S} \left| GradMetric(X) - \frac{ModelMetric}{ModelMetric_{exp}} \cdot GradMetric_{exp} \cdot \mu \right|, \mu > 0
\end{equation}
Where $S$ are the gradients of samples from the \textit{selection strides} which form the current batch, $ModelMetric$ is the  \textit{model metric} for the current batch, $ModelMetric_{exp}$ and $GradMetric_{exp}$ are the exponential running means for the \textit{model metric} and the \textit{gradient metric} and the $\mu$ is a parameter which allows us to control the slope and is $1.0$ by default.

The intuition behind this function is that we want the model metric and gradient metric to behave similarly so that the magnitude of the update step is proportional to the model loss for the current batch.
For example, supposing that the \textit{model metric} increases for the current batch due to the presence of several instances that were not classified correctly, we would like the selected samples from the batch to have a similar increase in \textit{gradient metric} and consequently in the update step. 
In the general case, the \textit{model metric} will decrease over time, therefore we want our \textit{gradient metric} to decrease, adjusting to the change in overall model performance . 

Another minimizing function which we used but achieved slightly worse results is the absolute value of the difference between the two metrics: $ |\alpha \cdot GradMetric(X) - ModelMetric|, \alpha > 0$. In our implementation  $\alpha$ is a scaling factor used to bring the two values to the same order of magnitude.
Minimizing this function follows the same principles as above, however it is more unstable because it does not take the past evolution into account and it is very sensitive to the choice of $\alpha$.

\subsection{Selection strategy}

In Eq. \ref{argmin} we try to select the best subset of \textit{selection strides} that minimizes our function. 
However, there are $2^{|S|}$ possible subsets and a search through all of them is prohibitively costly.
Therefore, we have designed two greedy selection strategies.
While they have a lower complexity we compromise on the fact that they might not able to always find the best solution.

The first strategy is \textit{bottom up selection} and the pseudo-code can be seen at Algorithm \ref{bottom-up}, while the second strategy is \textit{top down selection} and the pseudo-code can be seen at Algorithm \ref{top-down}.

\begin{minipage}[t]{0.5\textwidth}
    \begin{algorithm}[H]
    \small
    \caption{Bottom up selection}\label{bottom-up}
    \DontPrintSemicolon
      \SetKwFunction{FMain}{BottomUpSelection}
      \SetKwProg{Fn}{Function}{:}{}
      \Fn{\FMain{$selectionStrides$}}{
            $continue \gets true$ \\
            $best \gets \infty$ \\
            $chosen \gets \emptyset $ \\
            $m1 \gets ModelMetric()$ \\
            $m2 \gets ModelMetricMean()$\\
            \While{$continue$}{
                 $continue \gets false$ \\
                 \For{$stride \in selectionStrides$}{
                    $chosen \gets chosen \cup \{stride\}$ \\
                    $g1 \gets GradMetric(chosen)$ \\
                    $g2 \gets GradMetricMean(chosen)$\\
                    $d \gets F(m1, m2, g1, g2)$\\
                    \eIf{$d < best$}{
                        $best \gets d$\\
                        $selectionStrides \gets selectionStrides \setminus \{stride\}$\\
                        $continue \gets true$\\
                    }{
                        $chosen \gets chosen \setminus \{stride\}$
                    }
                 }
            }
            \KwRet $chosen$\;
      }
    \end{algorithm}
\end{minipage}
\hfill
\begin{minipage}[t]{0.5\textwidth}
    \begin{algorithm}[H]
    \small
    \caption{Top down selection}\label{top-down}
    \DontPrintSemicolon
      \SetKwFunction{FMain}{TopDownSelection}
      \SetKwProg{Fn}{Function}{:}{}
      \Fn{\FMain{$selectionStrides$}}{
            $continue \gets true$ \\
            $best \gets \infty$ \\
            $chosen \gets $selectionStrides$ $ \\
            $m1 \gets ModelMetric()$ \\
            $m2 \gets ModelMetricMean()$\\
            \While{$continue$}{
                 $continue \gets false$ \\
                 \For{$stride \in chosen$}{
                    $chosen \gets chosen \setminus \{stride\}$ \\
                    $g1 \gets GradMetric(chosen)$ \\
                    $g2 \gets GradMetricMean(chosen)$\\
                    $d \gets F(m1, m2, g1, g2)$\\
                    \eIf{$d < best$}{
                        $best \gets d$ \\
                        $continue \gets true$\\
                    }{
                        $chosen \gets chosen \cup \{stride\}$
                    }
                 }
            }
            \KwRet $chosen$\;
      }
    \end{algorithm}
\end{minipage}

Both strategies have a strong bias on the number of samples that would be included in an update.
\textit{Bottom up selection} usually selects few \textit{selection strides} while \textit{top down selection} selects more. 
As a consequence, training a model and using the \textit{top down selection} results in higher batch sizes than the \textit{bottom up selection}.
In order to combat this bias, when applying our algorithm for each layer of the neural network we randomly choose one of the two possible strategies with equal probability.

\subsection{Updating the batch size}

At the end of each epoch our optimizer calculates a batch size suitable for the next epoch.
We record how many samples were selected to be used in updates by our algorithm, and retrieve the 50th percentile, $q$, for these values. Using this, we calculate the batch size for the next training epoch to be:

\begin{equation}
\label{batch-change}
    Next = Current + 
    \begin{cases}
    delta, q > 0.8 \cdot Current\\
    -delta, q < 0.2 \cdot Current\\
    0, otherwise
    \end{cases}
\end{equation}

Where $Next$ represents the next batch size, $Current$ the current batch size and $delta$ a parameter with which we can control the difference between the current batch size and the next. 
The batch size is also capped at both ends by a minimum and maximum value, also given as parameters. 
The maximum batch size depends on hardware limitation and the size of the dataset, while the minimum batch size is needed to limit the maximum number of steps per epoch.

Putting it all together, our main training loop can be seen described in Algorithm~\ref{alg:train}.

\begin{algorithm}
\caption{Training process}\label{alg:train}
$dba \gets DBA(model, optimizer, strideSize, minBatchSize, maxBatchSize)$ \\
\While{$epochs < maxEpochs$}{
    $batches \gets SplitInBatches(data, batchSize)$ \\
    \For{$batch \in batches$}{
        $loss = model.fit(batch)$ \\
        $dba.step(loss)$
    }
    $batchSize \gets dba.nextBatchSize()$
}
\end{algorithm}

\section{Experimental results}
\label{experiments}

\begin{table}
  \caption{Results on MNIST using $x\%$ of the data}
  \label{sample-table}
  \centering
  \begin{tabular}{lllll}
    \toprule
    \multicolumn{2}{c}{} &
    \multicolumn{3}{c}{Test Accuracy \%}                   \\
    \cmidrule(r){3-5}
    Optimizer     & Variant     & 100\% of Data & 10\% of Data & 1\% of Data \\
    \midrule
    SGD & batch size=64  & $\textbf{98.164} \boldsymbol{\pm} \textbf{0.077}$ & $95.148 \pm 0.128$  & $87.628 \pm 0.339$\\
    SGD     & batch size=128 & $98.148 \pm 0.087$ & $94.990 \pm 0.076$  & $87.574 \pm 0.401$      \\
    SGD     & batch size=256       & $98.068 \pm 0.036$ & $94.834 \pm 0.137$  & $87.380 \pm 0.386$  \\
    Adam & batch size=64  & $97.786 \pm 0.037$ & $95.262 \pm 0.108$  & $88.350 \pm 0.330$   \\
    Adam     & batch size=128 & $97.770 \pm 0.068$ & $95.480 \pm 0.093$   & $88.250 \pm 0.452$  \\
    Adam     & batch size=256       & $97.872 \pm 0.085$ & $95.466 \pm 0.180$ & $88.314 \pm 0.545$ \\
    DBA+SGD \textbf{(ours)}    &  gradient norm      & $98.122 \pm 0.074$ & $95.546 \pm 0.073$  & $96.772 \pm 0.420$ \\
    DBA+SGD  \textbf{(ours)}   &  variance norm      & $98.140  \pm  0.032$ & $\textbf{97.422} \boldsymbol{\pm} \textbf{0.198}$ & $\textbf{97.830} \boldsymbol{\pm} \textbf{0.079}$ \\
    DBA+Adam  \textbf{(ours)}    &  gradient norm      & $95.038 \pm 0.118$ & $93.730 \pm 0.233$ & $90.846 \pm 1.749$ \\
    DBA+Adam  \textbf{(ours)}   &  variance norm      & $94.952 \pm 0.181$ & $93.864 \pm 0.214$ & $92.198 \pm 0.550$  \\
    \bottomrule
  \end{tabular}
\end{table}

We evaluated our method on random subsets of the MNIST Dataset \cite{lecun-mnisthandwrittendigit-2010},  ranging from $10$ samples per class to $6000$ samples per class which represents the entire dataset. The dataset is made available under the terms of the Creative Commons Attribution-Share Alike 3.0 license.
Each experiment was run five times using different random seeds and no data augmentation techniques are applied across the experiments.
The model we used is a Multilayer Perceptron with a single hidden layer of $64$ neurons.

For Stochastic Gradient Descent (SGD), we use a learning rate of $0.01$, weight decay of $0.0005$ and a Nesterov momentum of $0.9$. 
For Adam we use the same learning rate of $0.01$, while the beta coefficients used for computing running averages of gradient and its square are $0.9$ and $0.999$. 
The term added to the denominator to improve numerical stability, $\epsilon$ is $1\mathrm{e}{-8}$ and we use the same weight decay of $0.0005$. For both of these optimizers, a learning rate scheduler reduces the learning rate with a factor of $0.5$ until a minimum learning rate of $1\mathrm{e}{-7}$ is reached. The scheduler is activated when encountering a plateau and the training loss does not decrease by more than $0.05\%$ in 25 epochs. 

DBA is used as a wrapper for the previously mentioned optimizers. 
The minimum and maximum allowed batch sizes are 32 and $\min(|Data|, 2048)$ respectively. 
The \textit{selection stride} size is 16 for all the experiments, except for when we used 10 samples per class in which case we used \textit{selection strides} of size 10. 
When computing our exponential running average for the \textit{gradient metric} and the \textit{model metric} we used a smoothing factor of $0.9$ and the coefficient $\mu$ for the margin-based loss used when computing the slope for our metrics is $1.0$.
We used a parameter in order to specify which of the two possible \textit{gradient metrics} mentioned in Eq. ~\ref{grad-metric} are used.
The last parameter for DBA is the $delta$ we use to modify the batch size according to Eq. ~\ref{batch-change}, which is set to $8$ across our experiments.

For the practical implementation we used PyTorch \cite{NEURIPS2019_9015} which is publicly available under a modified BSD license. 
Due to the fact that the individual gradients in a batch are not easily available in the provided implementation, we used the open source library Opacus \cite{yousefpour2021opacus}, released under the Apache License 2.0, in order to recalculate and use them in our optimizer. 
This brings considerable additional overhead and a better implementation that parallelizes our metric calculations and selection algorithm would increase performance, however a complete discussion of these technical issues are beyond the scope of this paper. 
The training was done using two separate GPUs (RTX 1050 Ti and RTX 2080 Ti), each experiment being run on only one of them at a time.

Results for our experiments can be seen in Table~\ref{sample-table}. We include the average maximum accuracy and standard deviation obtained on the test split after 5 runs using different seeds. Results are reported for models trained using 100\%, 10\% and 1\% of the training data. The baseline training runs using SGD and Adam were repeated for different batch sizes, namely 64, 128 and 256 while the DBA runs start from a default 128 batch size. We also report the results of DBA runs using either gradient norm or variance norm as the \textit{gradient metric}.

\begin{figure}[h]
    \centering
    \begin{tikzpicture}[scale=0.90] 
        \begin{axis} [
                    anchor=east,
                    xshift=-1.7cm,
            x tick label style={font=\tiny},
            y tick label style={font=\tiny},
        symbolic x  coords={SGD+bs=25,SGD+bs=50,SGD+bs=100},xtick=data]
        \addplot+[forget plot,only marks, color={rgb,255:red,109; green,158; blue,235}] 
          plot[error bars/.cd, y dir=both, y explicit]
          table[x=x,y=y,y error plus expr=\thisrow{y-max},y error minus expr=\thisrow{y-min}] {\sgddata};
        \end{axis} 
        \begin{axis} [
                    anchor=west,
            x tick label style={font=\tiny},
            y tick label style={font=\tiny},
        symbolic x  coords={Adam+bs=25,Adam+bs=50,Adam+bs=100},xtick=data]
        \addplot+[forget plot,only marks, color={rgb,255:red,109; green,158; blue,235}] 
          plot[error bars/.cd, y dir=both, y explicit]
          table[x=x,y=y,y error plus expr=\thisrow{y-max},y error minus expr=\thisrow{y-min}] {\adamdata};
        \end{axis}
    \end{tikzpicture}
    \\
    \begin{tikzpicture}[scale=0.90] 
        \begin{axis} [
                    anchor=east,
                    xshift=-1.7cm,
            x tick label style={font=\tiny},
            y tick label style={font=\tiny},
        symbolic x  coords={DBA+SGD+VarianceNorm,DBA+SGD+GradientNorm},xtick=data]
        \addplot+[forget plot,only marks, color={rgb,255:red,109; green,158; blue,235}]  
          plot[error bars/.cd, y dir=both, y explicit]
          table[x=x,y=y,y error plus expr=\thisrow{y-max},y error minus expr=\thisrow{y-min}] {\dbasgddata};
        \end{axis} 
        
        \begin{axis} [
                    anchor=west,
            x tick label style={font=\tiny},
            y tick label style={font=\tiny},
        symbolic x  coords={DBA+Adam+VarianceNorm,DBA+Adam+GradientNorm},xtick=data]
        \addplot+[forget plot,only marks, color={rgb,255:red,109; green,158; blue,235}]  
          plot[error bars/.cd, y dir=both, y explicit]
          table[x=x,y=y,y error plus expr=\thisrow{y-max},y error minus expr=\thisrow{y-min}] {\dbaadamdata};
        \end{axis}
    \end{tikzpicture}
    
    \caption{Test error rates (\%) on MNIST after training using 10 samples per class}
    \label{fig:results_MNIST_10}
\end{figure}
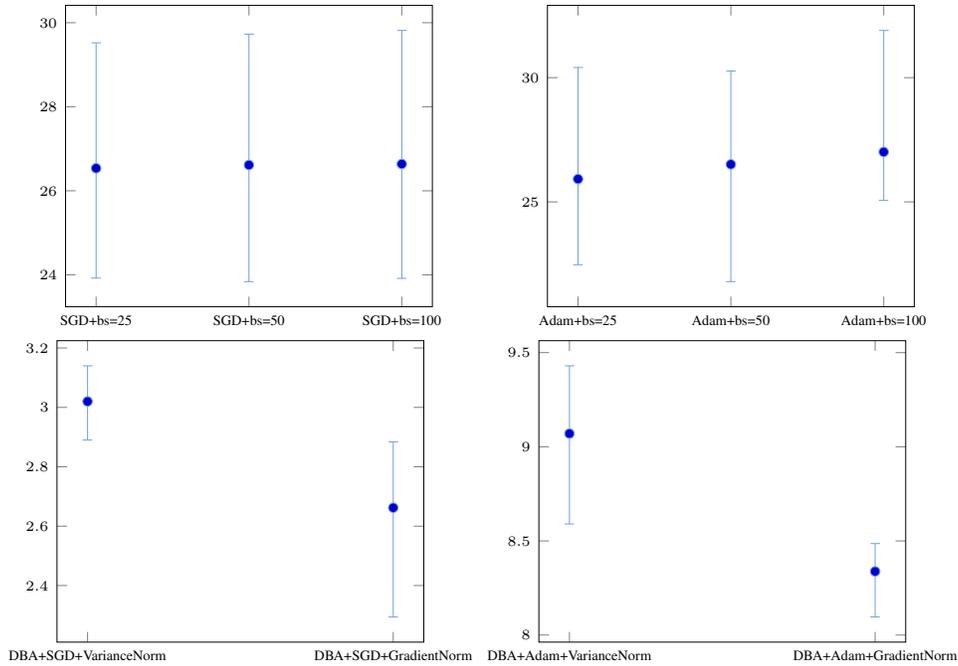

The results show that DBA performs extremely well in data-scarce environments, compared to SGD and Adam.
Interestingly enough, we have found out from the classic optimizers, Adam outperforms SGD when using less samples, showing that Adam's per-weight learning rate acts better with less data. 
However, the results for DBA + Adam are worse than DBA + SGD, and we suppose it is due to the fact that our selection of per-sample gradients interferes with Adam's exponential running mean mechanism.

When training on 10\% of the data (600 samples per class) our method achieved a lower accuracy than when training on 1\% of the data (60 samples per class). 
We weren't able to identify the reason for this phenomenon but we suspect that it's related to our choice for the \textit{selection stride} size, which is $16$ and might not be suitable for the 10\% dataset. 
Nonetheless, all the DBA+SGD combinations heavily outperform standard optimizers for the smaller subsets of MNIST with a significant difference being observed when training using only 1\% (60 samples per class). In this scenario one of our DBA+SGD runs using variance norm managed to reach an test accuracy of 97.79\% compared to the best performing baseline run (SGD with a batch size of 64) which managed to reach a maximum test accuracy of 87.87\%.
An important mention being that even if DBA didn't exceed certain baseline runs, it has performed comparably while it has the advantage of not needing to find a good batch size beforehand.

Motivated by our method's performance on small datasets, we also conducted experiments in an extreme data scarce scenario, namely using only 10 samples per class (100 samples total) of the MNIST dataset. We used the same parameters as mentioned above, except for the following:
\begin{itemize}
    \item Baseline runs now used batch sizes of 25, 50, 100 instead of 64, 128, 256;
    \item DBA runs use a starting batch size of 100 with a selection stride of 10 and minimum batch size of 20;
    \item Note that the maximum batch size is 100;
\end{itemize}

We report the test error rates in Figure~\ref{fig:results_MNIST_10}. As it can be seen, all DBA runs manage to significantly reduce the error rate compared to standard training. Notably, the lowest error rate achieved, 2.56\%, was the result of a model trained using DBA+SGD and gradient norm as the \textit{gradient metric}.

\begin{figure}[h]
    \centering
    \begin{tikzpicture}
        \centering
        \begin{axis}[anchor=east,
                    xshift=-1.7cm,
                    ylabel= Data Utilization Rate,
                    xlabel= Epochs,
                    ymin=0.0,
                    ymax=1.0,]
        \addplot[line width=1.pt, color={rgb,255:red,109; green,158; blue,235}]  table [x=Step, y=Value, col sep=comma, mark=none, smooth] {data/smooth_epoch_utilization.csv};
        \end{axis}
    
        \begin{axis}[anchor=west,
                    ylabel= "Real Epochs",
                    xlabel= Epochs,]
        \addplot[line width=1.pt, color={rgb,255:red,109; green,158; blue,235}]   table [x=Step, y=Value, col sep=comma, mark=none, smooth] {data/smooth_real_epoch_count.csv};
        \end{axis}
    \end{tikzpicture}
    \caption{Epoch utilization rate and "Real Epochs"}
    \label{fig:convergence_results}
\end{figure}
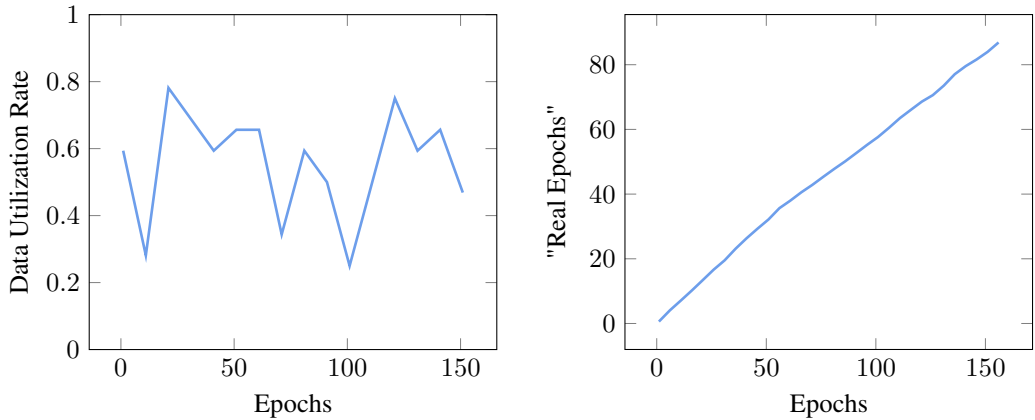

Models trained using DBA need to be trained on a relatively small volume of data to reach convergence, however, since our method discards many samples during training, this improvement in convergence is not reflected in the number of iteration used. In Figure~\ref{fig:convergence_results} we present the average ratio between number of selected gradients to the batch size for one of our DBA+SGD runs on the complete dataset. The data utilization rate indicates that only $\sim56\%$ of the data passing thought the model gets included in the update steps. This is cumulatively represented in the right side figure in a metric which we call "Real Epochs". It can be seen that after 150 epochs of training, the model only used $\sim83$ epochs worth of data for its updates.

\section{Limitations}
\label{limitations}

Regarding this work's limitations, we note that since layers get trained on different subsets of the current batch, issues caused by internal covariate shift might present themselves when using DBA on deeper models and this issue was not visible in our experiments.
Moreover we note that DBA is computationally expensive, due to the fact that it requires an additional per-sample gradients calculation and our selection procedure is sequential in our current implementation.
This can be mitigated by deriving better selection procedures and a lower level implementation for both features.

Furthermore, the experiments are not performed on a more difficult task and using deeper models, mainly because the generalization benefits of the DBA seem to taper off when we have enough data. 
Nevertheless, due to limited access to computational power we were not able to perform an extensive hyperparameter search.

\section{Conclusion}
\label{conclusion}
In this paper, we have introduced DBA, an algorithm that dynamically selects gradient samples for each layer to be included in weight updates.
The general consensus is that randomness in selecting samples plays a crucial role in ensuring a fast, stable and qualitative training process.
However, our results show that although we directly interfere with batch compositions, we manage to match the accuracy of standard training and even significantly exceed it when training in data scarce environments, such as using only 600 or 100 samples of the MNIST training dataset. 
This indicates that the quality of the gradients we select has a significant impact on the convergence speed and generalization capabilities. Moreover our approach is model and optimizer agnostic, which means it has the potential of being applied in many use-cases.

Future work should focus on developing more efficient ways to calculate per-sample gradients and designing selection methods in order to scale up such approaches to larger models.

{\small
\bibliographystyle{ieee_fullname}
\bibliography{main}
}

\end{document}